\documentclass[
    sigconf = true,     % MK: Style for GECCO proceedings (two columns and such).
    review = false,      % MK: Shows line number.
    screen = true,      % MK: Highlights links.
    anonymous = false,   % MK: Anonymous author information. (Also removes the Acknowledgment section.)
]
{acmart}

\usepackage{booktabs} % For formal tables
   \usepackage[flushleft]{threeparttable}
\usepackage{subfig}
\usepackage{multirow}
\usepackage{amsmath}
\setcopyright{none}
\usepackage{comment}
\usepackage{xcolor}

\newcommand{\tabitem}{~~\llap{\textbullet}~~}

\copyrightyear{2021} 
\acmYear{2021} 
\setcopyright{acmlicensed}\acmConference[GECCO '21]{2021 Genetic and Evolutionary Computation Conference}{July 10--14, 2021}{Lille, France}
\acmBooktitle{2021 Genetic and Evolutionary Computation Conference (GECCO '21), July 10--14, 2021, Lille, France}
\acmPrice{15.00}
\acmDOI{10.1145/3449639.3459407}
\acmISBN{978-1-4503-8350-9/21/07}

\usepackage[]{caption}

\begin{document}
\title[Personalizing Performance Regression Models]{Personalizing Performance Regression Models\\ to Black-Box Optimization Problems}
% \title{Personalized Optimization Problem Ensembles for Automated Algorithm Performance Prediction}

%%% The submitted version for review should be ANONYMOUS
\author{Tome Eftimov}
\orcid{}
\affiliation{%
  \institution{Jo\v{z}ef Stefan Institute}
  \streetaddress{Jamova cesta 39}
  \city{Ljubljana} 
 \country{Slovenia}
}
\email{tome.eftimov@ijs.si}

\author{Anja Jankovic}
\orcid{}
\affiliation{%
  \institution{Sorbonne Universit\'e, LIP6}
  \streetaddress{}
  \city{Paris} 
  \country{France}}
\email{anja.jankovic@lip6.fr}

\author{Gorjan Popovski}
\orcid{}
\affiliation{%
  \institution{Jo\v{z}ef Stefan Institute}
  \streetaddress{Jamova cesta 39}
  \city{Ljubljana} 
 \country{Slovenia}
}
\email{gorjan.popovski@ijs.si}

\author{Carola Doerr} 
\orcid{}
\affiliation{%
  \institution{Sorbonne Universit\'e, CNRS, LIP6}
  \streetaddress{}
  \city{Paris} 
  \country{France}}
\email{carola.doerr@lip6.fr}

\author{Peter Koro\v{s}ec}
\orcid{}
\affiliation{%
  \institution{Jo\v{z}ef Stefan Institute}
  \streetaddress{Jamova cesta 39}
  \city{Ljubljana} 
 \country{Slovenia}
}
\email{peter.korosec@ijs.si}

\renewcommand{\shortauthors}{T. Eftimov et al.}

\begin{abstract}
Accurately predicting the performance of different optimization algorithms for previously unseen problem instances is crucial for high-performing algorithm selection and configuration techniques. In the context of numerical optimization, supervised regression approaches built on top of exploratory landscape analysis are becoming very popular. From the point of view of Machine Learning (ML), however, the approaches are often rather na\"ive, using default regression or classification techniques without proper investigation of the suitability of the ML tools. With this work, we bring to the attention of our community the possibility to personalize regression models to specific types of optimization problems. Instead of aiming for a single model that works well across a whole set of possibly diverse problems, our personalized regression approach acknowledges that different models may suite different types of problems. Going one step further, we also investigate the impact of selecting not a single regression model per problem, but personalized ensembles. We test our approach on predicting the performance of numerical optimization heuristics on the BBOB benchmark collection.  
\end{abstract}



%\keywords{ACM proceedings, \LaTeX, text tagging}

\maketitle

\section{Introduction}
Before solving a real-world optimization problem via evolutionary or similar methods, users need to decide which of the many possible algorithms to apply (the \emph{algorithm selection (AS) problem}) and how to set its parameters (\emph{algorithm configuration (AC)}). While classically having to base all decisions on personal experience and recommendations to tackle these problems, users can rely today on the support of powerful AutoML techniques, which take as input some descriptors of the problem (and possibly also the specific instance) at hand, and which then recommend one or several algorithm instances that the users can apply to their problems. At the heart of many of these AutoML techniques are regression or classification algorithms, which -- in one way or the other -- predict which algorithm (configuration) could be most suitable for the given task. 

In evolutionary computation, a particularly active research question concerns the development of \emph{landscape-aware} AutoML methods~\cite{mersmann2011exploratory,KerschkeT19,BelkhirDSS17,MunozSurvey15,DerbelLVAT19,JankovicD20}. Classifying as \emph{supervised learning} approaches in the broader ML context, the main idea of these methods is to extract useful properties of the problem (instance) at hand, and to use this information to predict which algorithm (configuration) will perform particularly well on it. Landscape-aware AS and AC are trained on  performance data, which has been collected through previous optimization tasks or via systematic benchmarking.

\begin{figure*}[t!]
\includegraphics[width=0.7\textwidth]{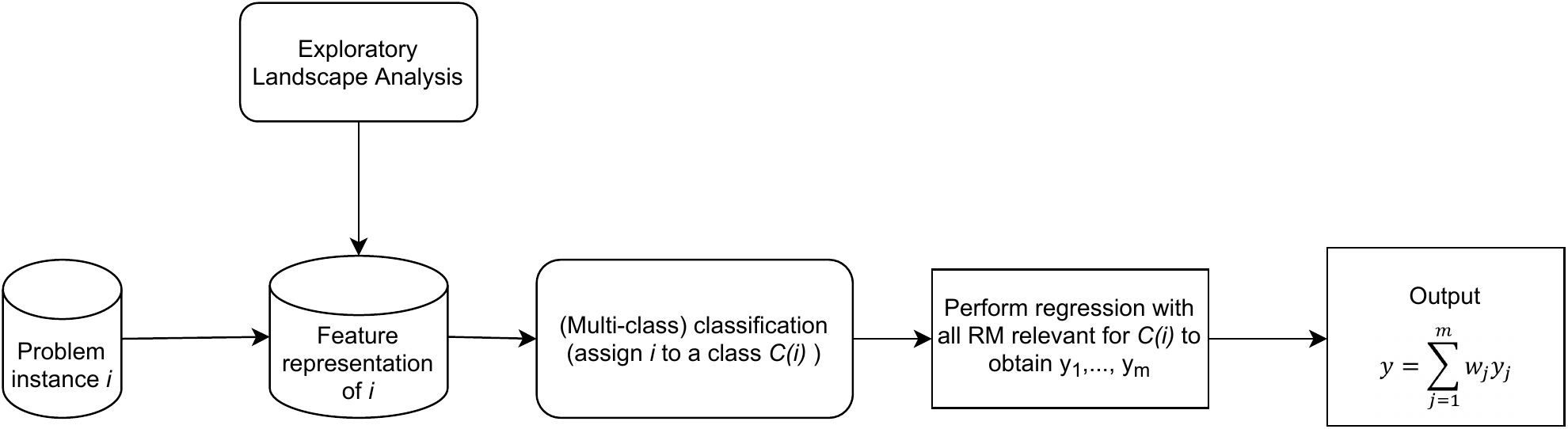}
\caption{Application of the personalized ML models. RM=regression models.} 
\label{fig:testing_phase}
\end{figure*}

Since the focus of most of the studies in evolutionary computation is on the development or assessment of different features, the Machine Learning (ML) model to derive the recommendation is often neglected, and authors satisfy themselves by applying off-the-shelf techniques such as default implementations of random forests (RF), support vector machines (SVM), or decision trees (DT), as, for example, available in the \texttt{scikit} toolbox~\cite{pedregosa2011scikit}. It is well known, however, that different predictive models can show quite different performances on different ML tasks. A more systematic approach towards the model selection seems therefore in order. 

\textbf{Our Contribution:} The main goal of our work is to analyze to what extent state-of-the-art landscape-aware AS and AC models could benefit from a more careful choice of the ML tools, and, more concretely, how their complementarity can be leveraged to obtain good predictions for broad sets of optimization problems. We evaluate a way of extending the current practice of deploying a single predictive model by combining ensemble regression and personalized regression. 
While ensemble learning is well-known and the de-facto standard in several ML-applications~\cite{zhang2012ensemble}, the idea to personalize the regressions is an original research contribution that we propose in this work. In a nutshell, the key idea is that different regression models work best for different types of problems, so that we can improve regression quality by automatically selecting the one(s) that showed best performance for similar problems.

We test the impact of each of the suggested extensions on a portfolio of 12 algorithms from the BBOB workshop series~\cite{BBOBdata}, for which task ourselves with predicting the solution quality after a fixed budget of function evaluations, a setting previously suggested in~\cite{JankovicD20}.

\textbf{Results:} We find that the regression quality improves for 58\%-70\% of the tested problems, depending on the comparison scenario, which nicely demonstrates that the current practice in performance regression used within the evolutionary computation community has quite some untapped potential.  

The computational overhead for training and applying the personalized models is also negligible for the tasks performed in this study. This happens because we are working only with 120 instances from 24 problems. When moving to larger data sets of several GB or even TB in size (as often considered in ML applications), however, we need to consider the impact of data size and data quality on the learning algorithms performances. With larger amounts of data, the so-called offline learning (as used in this paper) can become computationally inefficient. In such cases, one option is to use random or stratified sampling if possible, which can reduce the data size and still preserve the relevant information found in the original data set.

\textbf{Broader Impact of our Approach:} 
The idea to personalize the regression models is not restricted to performance regression, and not even to optimization. Based on our findings presented in this work, we consider further applications, for example in personalized medicine, as an exciting avenue for future work, since predictive models that are specifically developed for different genotypes and/or phenotype may allow better recommendations that one-size-fits all predictive models, which -- unfortunately -- are known to come with biases that can cause severe harm.

Note that personalized approach is different from training individual models separately, in that the classification which model to use is done in a data-driven way, and not by an external entity. While we use in this work classification to assign problem instances to problem classes, our approach can easily be extended to allow for interpolation between personalized models. How much such an additional layer would contribute, however, remains to be evaluated in future work. 
We compare our results to the approach in which the classification step can be omitted and the ground truth problem class is known (this is the ``Ensemble-ground'' method in Sec.~\ref{sec:setup}.) Based on the results, the misclassification of the problem class that leads to selecting a personalized model does not have a significant loss. Even more it can improve the end predictions, however this happens in cases where landscape representations of the problems instances are quite similar.   

\textbf{Structure of the Paper:} We summarize selected related works in Sec.~\ref{sec:relwork}. The pipeline to create the personalized regression models is presented in Sec.~\ref{sec:personalized}, and we discuss a specific use-case in Sec.~\ref{sec:usecase}. A critical discussion of our approach and avenues for future work are presented in Sec.~\ref{sec:conclusions}.      

\textbf{Availability of Data and Code:} All project data and code is available at~\cite{data}. Note that this repository does not only include data for the use-cases presented in Sec.~\ref{sec:usecase}, but also for all twelve algorithms from the selected portfolio. 

\section{Related Work}
\label{sec:relwork}
Our work integrates into ongoing research on landscape-aware algorithm selection (AS) and configuration (AC)~\cite{HutterKV19,LiefoogheDVDAT20,BlotMJH19,KerschkeT19,JankovicD20}, and more specifically to the task of per-instance algorithm selection (PIAS~\cite{KerschkeHNT19}) and configuration (PIAC~\cite{HutterHHL06PIAC,BelkhirDSS17}), which aims at recommending a best suited algorithm (configuration) based on the characteristics of the specific problem instance at hand. 

In landscape-aware AS/AC, recommendations are based on \emph{features} of the problem instances, which are estimated from a finite set $\{(x,f(x))\}$ of evaluated samples via so-called \emph{exploratory landscape analysis}~\cite{mersmann2011exploratory}. Most research in this area focuses on the definition or the analysis of features that describe certain characteristics of the optimization problem, and their suitability for automated algorithm selection, configuration, and design. See~\cite{MunozKH15InformationContent,LunacekW06dispersion,mersmann2011exploratory} for examples and further references. 

Where dynamic (``online'') algorithm configuration is considered, recommendations can also be based on the search behavior or algorithms' state parameters, see~\cite{DerbelLVAT19,BajerPRH19} for examples and further references. 

Concerning the ML techniques, we already mentioned that ensemble learning is quite standard nowadays, as it can result into much better predictions that using only one predictive model~\cite{mendes2012ensemble,qiu2014ensemble}. Our idea to build personalized regression models is based on a recent study in personalized nutrition~\cite{ispirova2020p}, which showed that for different clusters of recipes, different regression models provided the best prediction of the macronutrient values. None of the single predictive model could achieve similar performance in that task. This inspired us to automate the model selection in the context of landscape-aware AS/AC.     

\begin{figure*}[t!]
\includegraphics[width=0.75\textwidth]{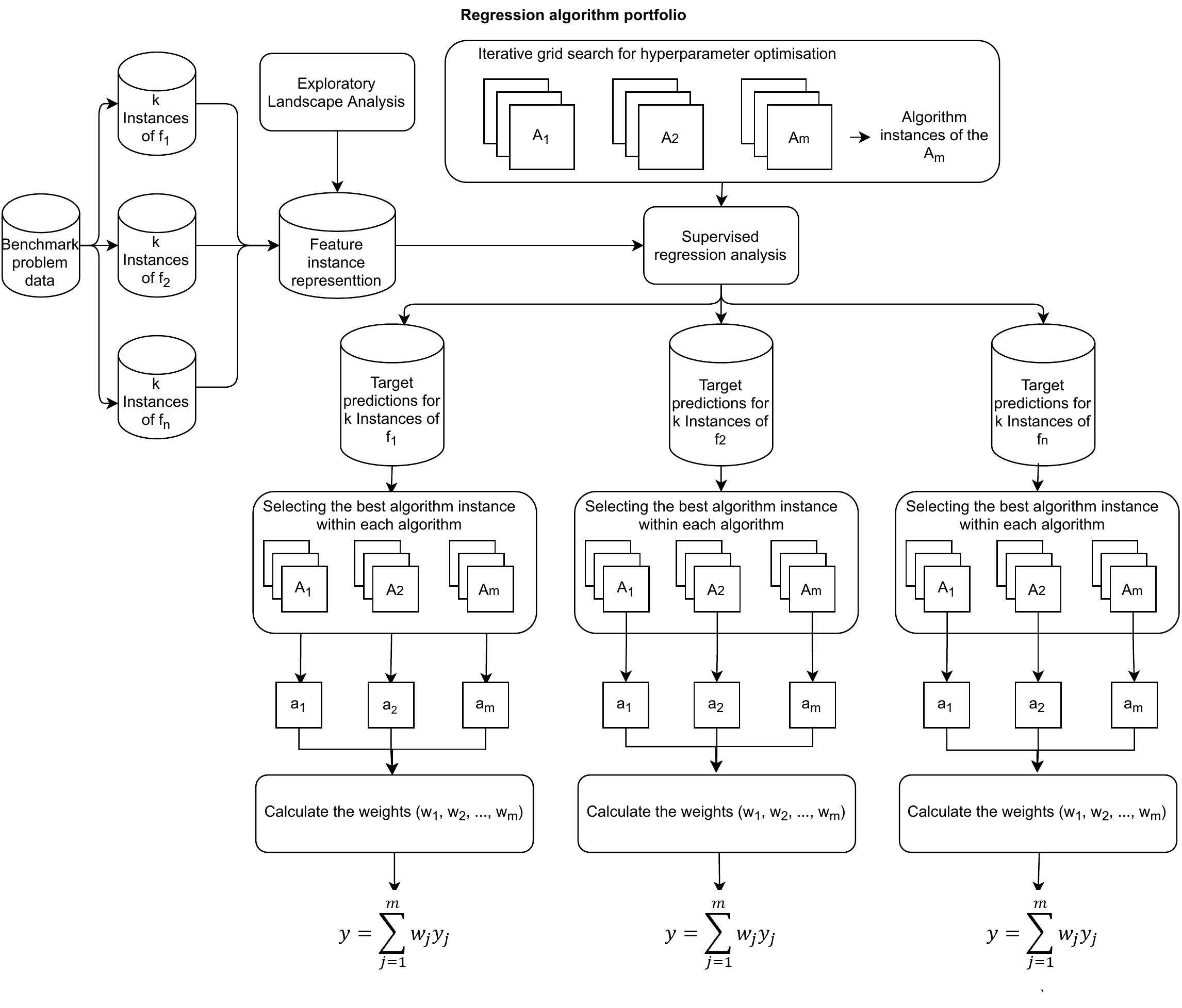} 
\caption{Training phase of the personalized problem ensembles.}
\label{fig:training_phase}
\end{figure*}

%%%%%%%%%%%%%%%%%%%%%%%%%%%%%%%%%%%%%%%%%%%%%%%%%%%%%%%%%%%
%%%%%%%%%%%%%%%%%%% SEC 2 %%%%%%%%%%%%%%%%%%%%%%%%%%%%%%%%%
\section{Personalized ML Models}
% \section{Personalized Performance Regression Models}
\label{sec:personalized}

To introduce our personalized performance regression pipeline, let us assume that we face a fixed-budget performance regression task, i.e., we assume to be aiming at predicting algorithms' solution quality after a fixed budget of function evaluations has been exhausted. The pipeline presented below can be used for other ML tasks, but the restriction to a specific use case eases the presentation considerably. 
 
The high-level approach is depicted in Fig.~\ref{fig:testing_phase}. The four main steps to obtain a performance prediction $y$ for a given problem instance~$i$ are as follows: 
\begin{enumerate}
    \item We first apply a feature extraction method to obtain a description of this instance $i$. In our context, the features are computed with exploratory landscape analysis (see Sec.~\ref{sec:usecase} for details).
    \item  We use the instance description to assign instance $i$ to a class $C(i)$ (i.e., we perform multi-class classification) to obtain a set $\{y_1, \ldots, y_m\}$ of different performance predictions, one per regression model (RM).
    \item Combine these values to the final prediction $y=\sum_{j=1}^m{w_j(i) y_j}$, using the weighting scheme $w_1(i), \ldots, w_m(i)$ of class $C(i)$.% derived in the last step of the training phase.
\end{enumerate}

The association of the RMs to the different classes as well as the computation of the class-specific weights is handled in a prior (``offline'') \textbf{training phase}. Its most relevant steps are illustrated in Fig.~\ref{fig:training_phase}.

Assuming that we have 
a set of training instances which are grouped into $n$ classes $C_1, \ldots, C_n$ (in our case, these are the problem instances), a set of potential RMs, which are grouped in to $m$ classes $A_1, \ldots, A_m$,\footnote{In this work, we group in one class all RMs that differ only in the hyper-parameters but use the same basic regression technique.}, 
and fixed-budget performance data for an algorithm $\mathcal{A}$, 
the training phase comprises the following steps: 
\begin{enumerate}
    \item Compute for each training instance a representation, ideally using the same feature extraction technique that will be used in the applications (i.e., in the ``test phase'' in proper ML terminology).
    \item Each RM instance uses the problem representation and the algorithm performance data to train a predictive model. 
    \item Each RM is evaluated according to its regression performance on the train instances within each optimization problem. 
    \item For each problem class $C(i)$, we select from each RM class $A_j$ the configuration $a_j(i)$ which achieved the best performance. 
    \item We then calculate the importance of each RM $a_j(i)$ via a min-max normalization. That is, if we denote by $q(i)=(q_1(i), \ldots, q_m(i))$ the vector of performance measures for each of the $m$ selected configurations for class $C(i)$, the importance of $a_j(i)$ is computed as 
    \begin{equation}
    q_{j,\text{norm.}}(i)=\frac{\max(q(i))-q_j(i)}{\max(q(i))-\min(q(i))},
    \end{equation}
    where  we assume a performance measure for which lower values are better (typically, deviation from the ground truth is measured in one way or the other). We then compute the vector $w(i)=(w_1(i), \ldots, w_m(i))$ of weights $w_j=\frac{q_{j,\text{norm.}}(i)}{\sum_{j=1}^m q_{j,\text{norm.}}(i)}$, which are used in the fourth step of the application phase described above. 
\end{enumerate}

%%%%%%%%%%%%%%%%%%%%%%%%%%%%%%%%%%%%%%%%%%%%%%%%%%%%%%%%%%%
%%%%%%%%%%%%%%%%%%% SEC 3 %%%%%%%%%%%%%%%%%%%%%%%%%%%%%%%%%

\section{Use-Case: ELA-based Fixed-Budget Performance Regression}
\label{sec:usecase}

We evaluate our personalized ML pipeline on a standard regression task which aims at predicting the final solution quality of a black-box optimization algorithm after a fixed number of function evaluations. The experimental setup is described in Sec.~\ref{sec:setup}. In total, we apply our approach to twelve different optimization algorithms. As a consequence of space limitations, however, we present here only some selected results (Sec.~\ref{sec:results}). A few sensitivity analyses, to test the robustness of our approach, are performed in Sec.~\ref{sec:sensitivity}.     

%%%%%
%%%%%
\subsection{Experimental Setup}
\label{sec:setup}

The experiments were performed on 13-inch MacBook Pro with 2.8 GHz Quad-Core Intel Core i7 processor and 16 GB of RAM. The raw regression data has been collected using Python implementation, while the personalized approach and all the evaluations have been performed using R. Our fixed-budget regression is inspired by~\cite{JankovicD20}, but applied here to the more diverse set of algorithms suggested in~\cite{KerschkeT19}. Concretely, we aim at predicting the performance of the following 12 algorithms: 
BrentSTEPqi~\cite{PosikB15}, BrentSTEPrr~\cite{PosikB15}, CMA-ES-CSA~\cite{atamna2015benchmarking}, HCMA~\cite{loshchilov2013bi}, HMLSL~\cite{pal2013benchmarking}, IPOP400D~\cite{auger2013benchmarking}, MCS~\cite{huyer2009benchmarking}, MLSL~\cite{pal2013benchmarking}, OQNLP~\cite{pal2013comparison}, fmincon~\cite{pal2013comparison}, fminunc~\cite{pal2013comparison}, and BIPOP-CMA-ES~\cite{hansen2009benchmarking}. Note here that the latter does not appear in the portfolio analyzed in~\cite{KerschkeT19}, but it was added since for one algorithm from the original study the raw performance data was missing.
As performance measure of these algorithms we use their single run fixed-budget target precision after 250, 500, and 1\,000 fitness evaluations, respectively, and this for the first five instances of each of the 24 BBOB functions~\cite{bbobfunctions} provided by the BBOB platform~\cite{hansen2020coco}. This performance data is available at~\cite{BBOBdata}, but for our work we used the post-processed and more conveniently queryable repository available at~\cite{IOHanalyzer}.  The representations of the 120 problem instances are based on exploratory landscape analysis~\cite{mersmann2011exploratory}. The features were computed by the R-package \texttt{flacco}~\cite{flacco}, using the uniform sampling procedure with a budget of $400d$ (a sensitivity analysis for a $50d$ sampling budget will be presented in Sec.~\ref{sec:sensitivity}). 
Following~\cite{JankovicD20} we used 56 feature values per instance, which are grouped into five features groups: disp, ela\_level, ela\_meta, ic, and nbc (see~\cite{JankovicD20} for full names and references). To stick to common practice in the evolutionary computation community, we take the raw feature values, i.e., we do not normalize these values nor do we perform any representation learning prior to feeding the values to our ML approaches.  

To evaluate the personalized ensembles, we used stratified 5-cross fold validation, where each fold consists of the first, second, third, fourth, and fifth instances for each problem, respectively. That is, we repeat the whole training and testing process described in Sec.~\ref{sec:personalized} five times, each time leaving out one fold for the testing phase and using the other four for the training. Note that the personalized ensembles for the same problem can be different across the five different runs, since different training data is used. An example will be presented in Table~\ref{tab:models}. We have taken the stratified 5-cross fold validation approach, since it is the predominantly used one in evolutionary computation~\cite{DerbelLVAT19,JankovicD20}.

In a first evaluation of our personalization approach, we used seven regression techniques: Lasso~\cite{tibshirani1996lasso}, ElasticNet~\cite{zou2005regularization}, KernelRidge~\cite{murphy2012machine}, PassiveAggressive~\cite{crammer2006online}, DecisionTree~\cite{breiman1984classification}, RandomForest~\cite{breiman2001random}, and BaggingDT~\cite{breiman1996bagging}. 
We applied iterative grid search to each of them to test different hyper-parameters. Evaluating the seven regression techniques using the mean absolute errors (MAE) of the test folds from the stratified 5-cross fold validation, only three regression techniques were selected for further investigation, DecisionTree, RandomForest, and BaggingDT. The tested hyper-parameter combinations for each of these techniques are summarized in Table~\ref{tab:hyperparameters_regression}. Since all selected techniques are based on trees, the \textit{crit} parameter can be ``mse" - mean squared error, ``mae" - mean absolute error and ``friedman$\_$mse" - Friedman mean squared error. Regarding the \textit{minsplit} hyper-parameter, it is the minimum number of data instances a node contains in order to be slit. The \textit{nest} hyper-parameter defines how many Decision Trees will be built in the RandomForest/BaggingDT RM. These range of the hyper-parameters have been selected concerning the data set size and the guidelines available in ML to avoid overfitting.  In total, we ended up with 430 different RMs, 30 configurations of DecisionTree,  200 configurations for RandomForest, and 200 configurations of BaggingDT. 
\setlength\abovecaptionskip{-0.5ex}
\begin{table}[tb]
    \centering
     \caption{Hyper-parameter values for each regression model class.}
    \resizebox{.4\textwidth}{!}{% <------ Don't forget this %
    \begin{tabular}{l|l}
    \hline
    Algorithm & Hyperparameters\\
    \hline
        DecisionTree & \tabitem  $crit \in \{"mse", "mae", "friedman\_mse"\}$ \\
         (30 configs.) & \tabitem  $minsplit \in \{2,4,6,8,10,12,14,16,18,20\}$ \\
         \hline
        RandomForest & \tabitem  $crit \in \{"mse", "mae"\}$ \\
         (200 configs.)& \tabitem  $minsplit \in \{2,4,6,8,10,12,14,16,18,20\}$ \\
         & \tabitem $nest \in \{10,20,30,40,50,60,70,80,90,100\}$ \\
        \hline
        BaggingDT & \tabitem  $crit \in \{"mse", "mae"\}$ \\
         (200 configs.)& \tabitem  $minsplit \in \{2,4,6,8,10,12,14,16,18,20\}$ \\
         & \tabitem $nest \in \{10,20,30,40,50,60,70,80,90,100\}$ \\

        \hline
    \end{tabular}% <------ Don't forget this %
   }
 
    \label{tab:hyperparameters_regression}
\end{table}

To select the best RM within the three selected regression techniques and to learn their weights for each problem class separately, the mean absolute error from the results obtained on the training fold has been used. This has been done since the data set we used is relatively small, and we cannot split it into train, validation, and test set. In the future, when working with much bigger data sets, validation sets should be used for the weight calculation. Currently, this can lead further to overfitting on the training set, however it can also provide preliminary information about how the methodology fits for new test problem instances.

To associate an instance $i$ to a problem class $C(i)$, an ensemble with majority vote of three multi-class classification algorithms (BaggingDT\_crit-entropy\_minsplit-2\_nest-9, RandomForest\_entro\-py\_nest-9\_min-2, and RandomForest\_gini\_nest-9\_min-2) was train\-ed on the same folds used for building the personalized regression models. The hyper-parameters used for training the classifiers are the same as the regression models, with the difference that Gini impurity (\textit{gini}) or the Information Gain (\textit{entropy})~\cite{friedman2001elements} were used for splitting the nodes in the individual tree. They both measure the impurity of a node.

\begin{table}[!tb]
\centering
\caption{Regression models used for evaluation purposes.}
\label{tab:models_comp}
\resizebox{.48\textwidth}{!}{% <------ Don't forget this %
\begin{tabular}{l|c|c|c|c|c}
\hline
\multicolumn{1}{c|}{\multirow{2}{*}{Regression   model}} &  &  & \multicolumn{2}{c|}{Selection}                                          &  \\

\multicolumn{1}{c|}{}                                    &          Personalized     &  Ensemble        & \multicolumn{1}{c|}{MAE   on train} & \multicolumn{1}{c|}{MAE   on test} &      Classification          \\
\hline
Ensemble-ground                                         &   \checkmark           &   \checkmark       &                   \checkmark                 &         -                         &      -          \\
Ensemble-class                                          &      \checkmark        &      \checkmark    &                      \checkmark              &                  -                 &          \checkmark      \\
Best-train                                              &      -        &     -    &      \checkmark                               &            -                       &        -        \\
Best-train-instance                                     &    \checkmark          &      -    &     \checkmark                               &         -                          &      -          \\
Best-test                                               &        -      &      -    &      -                              &                   \checkmark                 &    -        \\
\hline
\end{tabular}% <------ Don't forget this %
   }
\end{table}
The comparison is done in the following scenarios, which are summarized in Table~\ref{tab:models_comp}:
\begin{itemize}
    \item \textbf{Ensemble-ground}: personalized ensembles for each problem;  true problem class, $C(i)$, to which the test problem instance $i$ belongs, is known as a priori information. That is, we assume in this model that we know which problem class the instance belongs to and our key objective is to evaluate the appropriateness of the class-specific ensemble.
    \item \textbf{Ensemble-class}: this is the approach described in Sec.~\ref{sec:personalized}, i.e., we have personalized ensembles for each problem, and the problem class $C(i)$ has to be guessed from the instance representation by the classifier. When the classifier correctly predicts the truth problem class, the prediction will be identical to that of the Ensemble-ground model. When the instance is mis-classified to a different problem class, the noise presented in the classifier will affect the selection of the relevant RMs (i.e., RMs for different problem will be selected), which influences the end prediction (which can go in both ways, as we shall see below). 
    \item \textbf{Best-train}: the best RM from the three regression techniques is selected based on the MAE obtained  across all problems from the training folds (i.e., one RM for all problems).
   \item \textbf{Best-train-instance}: the best RM is selected in the same scenario as the Best-train, but for each problem separately (i.e., not across all problems). Selecting the Best-train-instance model for each problem is a special case of personalized models (i.e., each problem has its own best RM, but we do not combine the predictions of several model for the final output).
\item \textbf{Best-test}: the best RM from the three regression techniques is selected based on the MAE obtained  across all problems from the test folds (i.e., one RM for all problems).
\end{itemize}

Note that the first four above-mentioned models are learned without evaluating problem instances from the test folds, whereas testing is needed to select the Best-test RM.  
\begin{figure*}[th]
\includegraphics[width=0.8\textwidth]{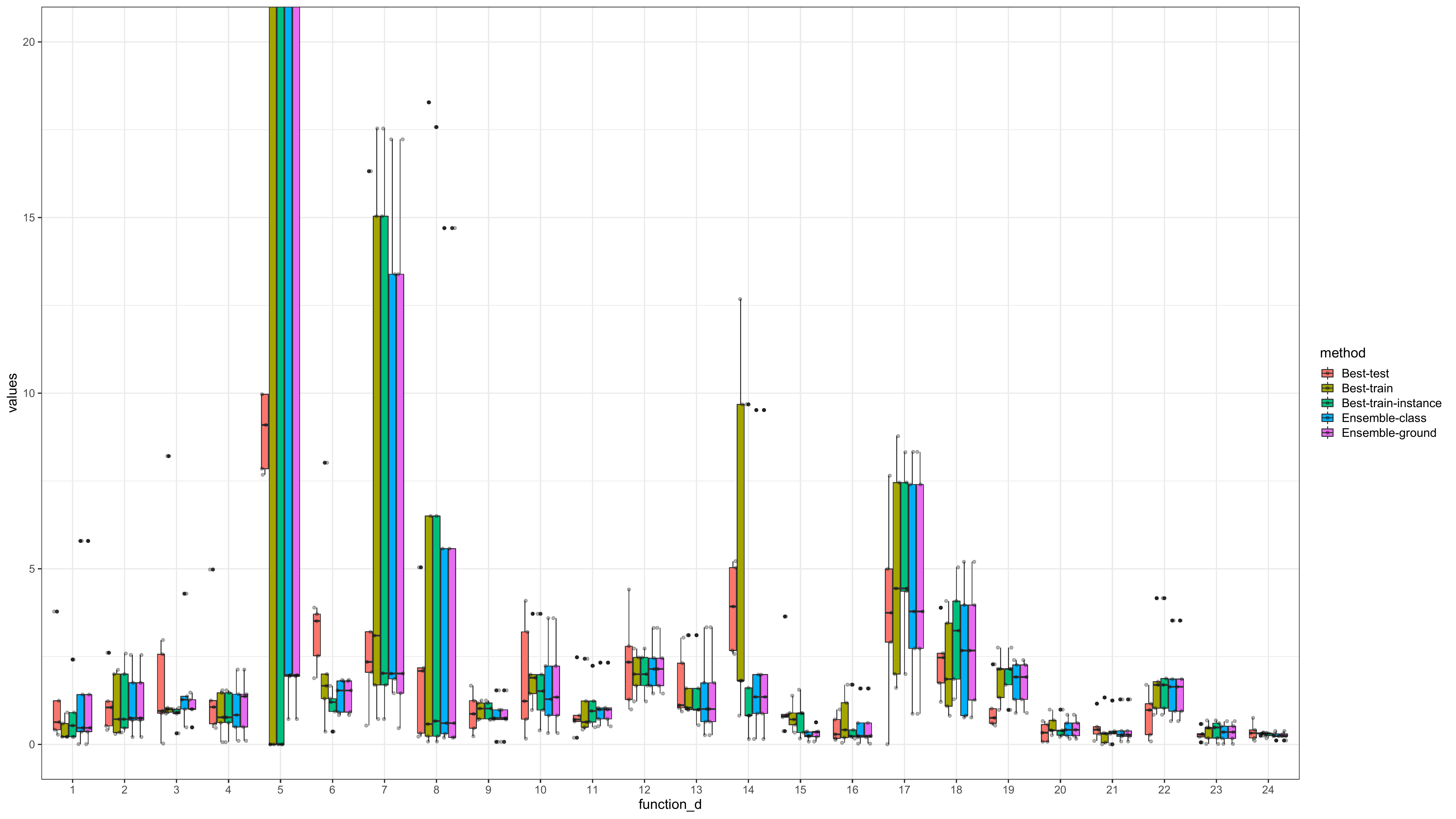}
\caption{Evaluation results for BIPOP-CMA-ES performance prediction. The $y$-axis corresponds to the absolute error between the truth and predicted target precision (i.e., natural logarithmic of the target precision), while the $x$-axis corresponds to each BBOB benchmark problem. The boxplots present the distribution of the absolute error obtained from each test fold for each problem separately in five different scenarios: i) Best-test, ii) Best-train, iii) Best-train-instance iv) Ensemble-class, and v) Ensemble-ground. } 
\label{fig:first}
\end{figure*}

\subsection{BIPOP-CMA-ES Performance Prediction}
\label{sec:results} 

To evaluate the proposed methodology, the scenario of BIPOP-CMA-ES performance prediction is explored. The experiment is performed for a fixed budget of 1000 function/solution evaluations, and an ELA feature portfolio calculated using a $400d$ sample size. The personalized ensembles were trained in two scenarios: once to predict the original target precision achieved, and once to predict the natural logarithm value of the target precision. Regardless of how the target is represented, the benefits of using the proposed methodology is the same. Since this paper is required to adhere to a page limit, we present the models with the natural logarithm value of the target precision. 

Figure~\ref{fig:first} shows the distribution of the absolute error obtained from each test fold for each problem separately in five different scenarios: i) Best-test, ii) Best-train, iii) Best-train-instance iv) Ensemble-class, and v) Ensemble-ground. The Best-test model is RandomFore\-st\_crit-mse\_minsplit-6\_nest-20,  while the Best-train model is Deci\-sionTree\_crit-mae\_minsplit-4. 
Table~\ref{tab:models} presents the models used in the personalized ensembles for the sixth problem (i.e., Attractive Sector Function) for each fold. 

\begin{table}[tb]
\centering
\caption{Regression models used in the personalized ensembles for the sixth problem in each fold.}
\label{tab:models}
\resizebox{.3\textwidth}{!}{% <------ Don't forget this %
\begin{tabular}{c|l}
  \hline
 Fold & Models \\ 
  \hline
1 & DecisionTree\_crit.mse\_minsplit.4\\
& RandomForest\_crit.mae\_minsplit.2\_nest.90\\
& BaggingDT\_crit.mae\_minsplit.2\_nest.10\\ 
\hline
  2 & DecisionTree\_crit.mae\_minsplit.4	\\
& RandomForest\_crit.mse\_minsplit.6\_nest.90\\
& BaggingDT\_crit.mae\_minsplit.6\_nest.10\\
\hline
 3 & DecisionTree\_crit.mse\_minsplit.4 \\
&	RandomForest\_crit.mse\_minsplit.2\_nest.20\\
& BaggingDT\_crit.mae\_minsplit.10\_nest.10\\
\hline
4 & DecisionTree\_crit.mae\_minsplit.4 \\
& RandomForest\_crit.mae\_minsplit.4\_nest.30\\
&BaggingDT\_crit.mae\_minsplit.2\_nest.10\\
\hline
  5 & DecisionTree\_crit.mae\_minsplit.4 \\
& RandomForest\_crit.mse\_minsplit.2\_nest.70\\
& BaggingDT\_crit.mse\_
minsplit.10\_nest.10\\
\hline
   \end{tabular} % <------ Don't forget this %
   }
\end{table}
 
 \textbf{Comparing a single RM  vs. personalized ensembles:} This comparison involves comparing the results obtained by using a single RM that works well across all problems (i.e., Best-train or Best-test) with the results obtained using the personalized ensembles (i.e., Ensemble-class and Ensemble-ground). 
 By comparing the Best-train, Ensemble-class, and Ensemble-ground models, looking in the medians from the boxplots, it is obvious that the personalized ensembles (i.e., Ensemble-class, and Ensemble-ground) are better than the Best-train for 14 out of 24 optimization problems (i.e., 6, 7, 9, 10, 13, 14, 15, 16, 17, 19, 21, 22, 23, and 24). This comparison actually involves models that have never seen the test instances. More promising results are actually obtained when comparing the  Ensemble-class and Ensemble-ground models with the Best-test model. In this case, the personalized ensembles are better in 15 out of 24 problems (i.e., 1, 2, 4, 5, 6, 7, 8, 9, 12, 13, 14, 15, 16, 21, and 24). In addition, we should point out that the selection of the Best-test is done using the information from the test instances, that have never been seen by the personalized models. Table~\ref{tab:median} presents the median absolute error obtained from each test fold for each problem separately by the four different regression models. Comparing the median absolute values we can see that the gain by using the personalized ensembles varies between the problems, but this also results from the different target precision range between the problems.

\textbf{Comparing the ground personalized ensembles with the personalized ensembles combined with classification:}
By comparing the median values between the Ensemble-class and Ensemble-ground models, we can actually see the influence of the classifier on the end result. When there is a difference between the end prediction results obtained by both models, it means that the classifier predicted the wrong problem class. This happens for four problems (i.e., 3, 4, 10, and 15). In the case of the fourth and the fifteenth problem, the misclassification actually improves the end target prediction. To see which RMs are selected and combined to generate the personalized ensemble, the confusion matrix from the classification is further explored. For the fourth problem, the misclassification happens in the third test fold, where the instance from the fourth problem class (i.e., Büche-Rastrigin function, which is a separable function), is assigned to the third problem (i.e., Rastrigin function, which is also a separable function). For the fifteenth problem (i.e., Rastrigin Function, which is a multi-modal function with adequate global structure), the misclassification happens in the first test fold, where the classifier classifies it into the third problem (i.e., Rastrigin Function, which is a separable function).

These results open new directions for future work; instead of training personalized ensembles on the problem level it will shift to learning them for a whole group of instances which belong to the same cluster. This cluster can be obtained by clustering the ELA representation of the problem instances.

\begin{table}[tb]
\centering
\caption{Median absolute error obtained from each test fold for each problem separately in four different scenarios.}
\label{tab:median}
\resizebox{.4\textwidth}{!}{% <------ Don't forget this %
\begin{threeparttable}
\begin{tabular}{c|cccc}
  \hline
 Problem & Best-test & Best-train & Ensemble-ground & Ensemble-class \\ 
  \hline
1 & 0.6337 & 0.2170 & 0.4718{$^\star$} & 0.4718{$^\star$} \\ 
  2 & 1.0507 & 0.7152 & 0.7478{$^\star$} & 0.7478{$^\star$} \\ 
  3 & 0.9530 & 0.9963 & 1.0053 & 1.2729 \\ 
  4 & 1.0627 & 0.7711 & 1.3661 & 0.8353{$^\star$} \\ 
  5 & 9.0958 & 0.0000 & 1.9736{$^\star$} & 1.9736{$^\star$} \\ 
  6 & 3.5115 & 1.6669 & 1.5341{$^\triangle$} & 1.5341{$^\triangle$} \\ 
  7 & 2.3472 & 3.0982 & 2.0179{$^\triangle$} & 2.0179{$^\triangle$} \\ 
  8 & 2.0920 & 0.5789 & 0.6052{$^\star$} & 0.6052{$^\star$} \\ 
  9 & 0.8675 & 1.0208 & 0.7480{$^\triangle$} & 0.7480{$^\triangle$} \\ 
  10 & 1.2298 & 1.8954 & 1.3431${^\diamond}$ & 1.2865$^{\diamond}$ \\ 
  11 & 0.7064 & 0.6347 & 0.9910 & 0.9910 \\ 
  12 & 2.3399 & 1.9997 & 2.1457{$^\star$} & 2.1457{$^\star$} \\ 
  13 & 1.1155 & 1.0443 & 1.0073${^\triangle}$ & 1.0073$^{\triangle}$ \\ 
  14 & 3.9245 & 1.8126 & 1.3522{$^\triangle$} & 1.3522{$^\triangle$} \\ 
  15 & 0.8055 & 0.7086 & 0.3540{$^\triangle$} & 0.2452$^{\triangle}$\\ 
  16 & 0.2879 & 0.4122 & 0.2404{$^\triangle$} & 0.2404$^{\triangle}$\\ 
  17 & 3.7476 & 4.4404 & 3.7838$^{\diamond}$ & 3.7838$^{\diamond}$ \\ 
  18 & 2.4728 & 1.8581 & 2.6707 & 2.6707 \\ 
  19 & 0.7567 & 2.1484 & 1.9176$^{\diamond}$ & 1.9176$^{\diamond}$ \\ 
  20 & 0.3334 & 0.4013 & 0.4151 & 0.4151 \\ 
  21 & 0.4166 & 0.3046 & 0.2726$^{\triangle}$ & 0.2726$^{\triangle}$ \\ 
  22 & 0.9757 & 1.6932 & 1.6386$^{\diamond}$ & 1.6386$^{\diamond}$ \\ 
  23 & 0.2916 & 0.4672 & 0.3494$^{\diamond}$ & 0.3494$^{\diamond}$ \\ 
  24 & 0.3209 & 0.3103 & 0.2335$^{\triangle}$ & 0.2335$^{\triangle}$ \\ 
   \hline
   \end{tabular} 
   \begin{tablenotes}
            \item[$\diamond$] Better than Best-train.
            \item[$\star$] Better than Best-test.
            \item[$\triangle$] Better than both, Best-train and Best-test.
        \end{tablenotes}
     \end{threeparttable}% <------ Don't forget this %
}
\end{table}

\textbf{Comparing a single personalized RM with personalized ensembles:}
To delve deeper, the Ensemble-class and Ensemble-ground personalized models are compared to the Best-train-instance model. In this scenario, we have limited it to only the best RM for each problem separately, learned using the performance obtained from the training folds, excluding the information from the test instances in the selection. Looking at the median absolute error across the test folds, the personalized ensembles are better than the Best-train-instance models for 13 out of 24 problems (i.e. the problems: 1, 7, 8, 9, 10, 15, 17, 18, 19, 21, 22, 23, and 24). Since the personalized ensembles are based on combining different RMs, the Best-train-instance model for each problem is actually one of the three RMs that are being combined. In most cases, combining the best RM (i.e., Best-train-instance) with the RMs from the other two regression techniques improves the final prediction. However, there are also cases when the end prediction is not better, which further opens the question of how to select the regression techniques that should be included in learning the ensembles.  

\textbf{Mean absolute error vs. median absolute error comparison:}
In the case when the mean absolute error is used for comparison, the personalized ensembles (i.e., Ensemble-class, and Ensemble-ground) are better than the Best-train in 17 out of 24 optimization problems (i.e., 3, 4, 5, 6, 7, 8, 9, 10, 13, 14, 15, 16, 17, 19, 20, 21, 22, 23, and 24). While comparing them with the Best-test model, they are better in 11 out of 24 problems (i.e., 3, 5, 6, 9, 10, 12, 13, 14, 15, 21, and 24). When the comparison is done with the Best-train-instance, the personalized ensembles are better in 17 out of 24 problems (i.e., 2, 5, 7, 8, 9, 10, 13, 15, 16, 17, 18, 19, 20, 21, 22, 23, and 24).

%%%%%%%%
%%%%%%%%
\subsection{Sensitivity Analysis}
\label{sec:sensitivity}

To investigate the impact of the different steps used by the methodology on the final predictions, we compare the Best-test, Best-train, and Ensemble-class models in three different scenarios:
\begin{enumerate}
    \item We investigate how different sample sizes required to calculate the ELA features influence the prediction of the reached target precision;
    \item Results for different budgets for one optimization algorithm and fixed ELA portfolio are presented;
    \item The results for different optimization algorithms are presented, for fixed ELA features portfolio and fixed budget.
\end{enumerate}

\begin{figure}
    \setlength{\belowcaptionskip}{-10pt}
    \centering
    \includegraphics[scale=0.33]{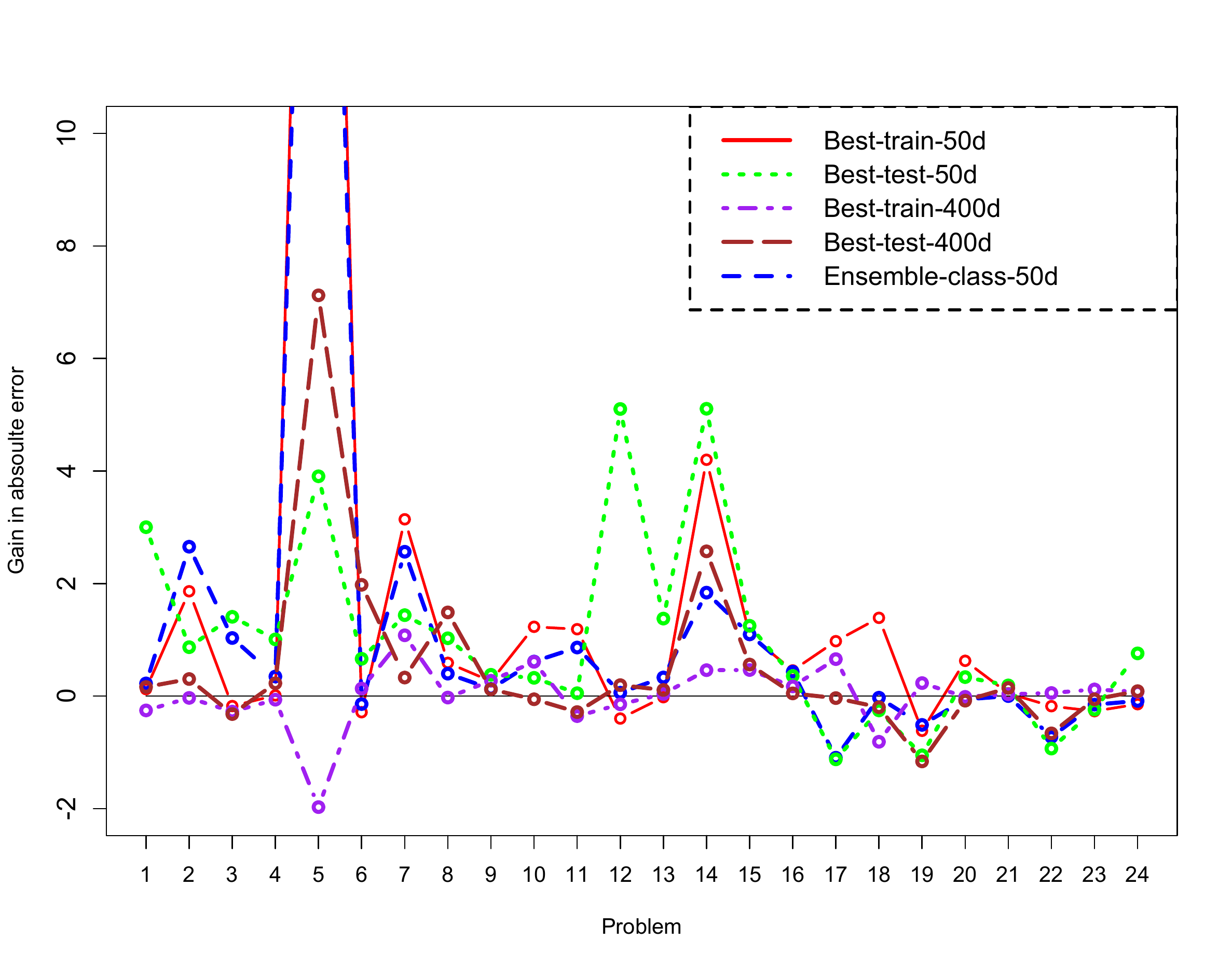}
    \caption{Relative advantage of the Ensemble-class-400d model vs. the regression models (Best-train-50d, Best-test-50d, Best-train-400d, Best-test-400d, Ensemble-class-50d). The y-axis presents the difference between the median absolute errors. Positive values indicate where the Ensemble-class-400d model is better than the other models, while negative values indicate vice-versa.}
    \label{fig:samplesize}
\end{figure}

\textbf{Fixed budget, one optimization algorithm, different sample sizes for ELA calculation:}
To investigate the impact of different sample sizes required to calculate the ELA features on the final predictions, the 56 selected ELA features were calculated using $50d$  and $400d$ sample sizes. This resulted in two ELA features portfolios, which were further used as input data to learn personalized ensembles for the BIPOP-CMA-ES in the fixed budget scenario, where the budget was set at 1000 evaluations. Figure~\ref{fig:samplesize} presents the relative advantage of the the Ensemble-class-400d model vs. the regression models (Best-train-50d, Best-test-50d, Best-train-400d, Best-test-400d, Ensemble-class-50d) for each problem separately. The suffix $50d$ or $400d$ in the name of each model presents which ELA feature portfolio is used for learning it. To estimate the advantage, the difference between the median absolute errors has been calculated. Positive values indicate where the Ensemble-class-400d model is better than the other models, while negative values indicate vice-versa. From the figure, it follows that the Ensemble-class-400d model is better than the other models in most of the problems. Comparing the Ensemble-class-50d and the Ensemble-class-400d (i.e., the blue line), it follows that using the ELA feature portfolio calculated with $400d$ sample size provides much better results in most of the problems, however there are few problems (i.e., 17, 19, 22) for which the opposite is true with really small differences in the median absolute errors. The results show us that there is a benefit of using the personalized ensembles. However the selection of the sample size required to calculate the ELA features can influence the final prediction. This also points to the robustness of the ELA features for different samples sizes, which has been already discussed and investigated from another perspective in a recently published paper~\cite{renau2020exploratory}. 

\begin{figure}[!tb]
    \setlength{\belowcaptionskip}{-10pt}
    \centering
    \includegraphics[scale=0.28]{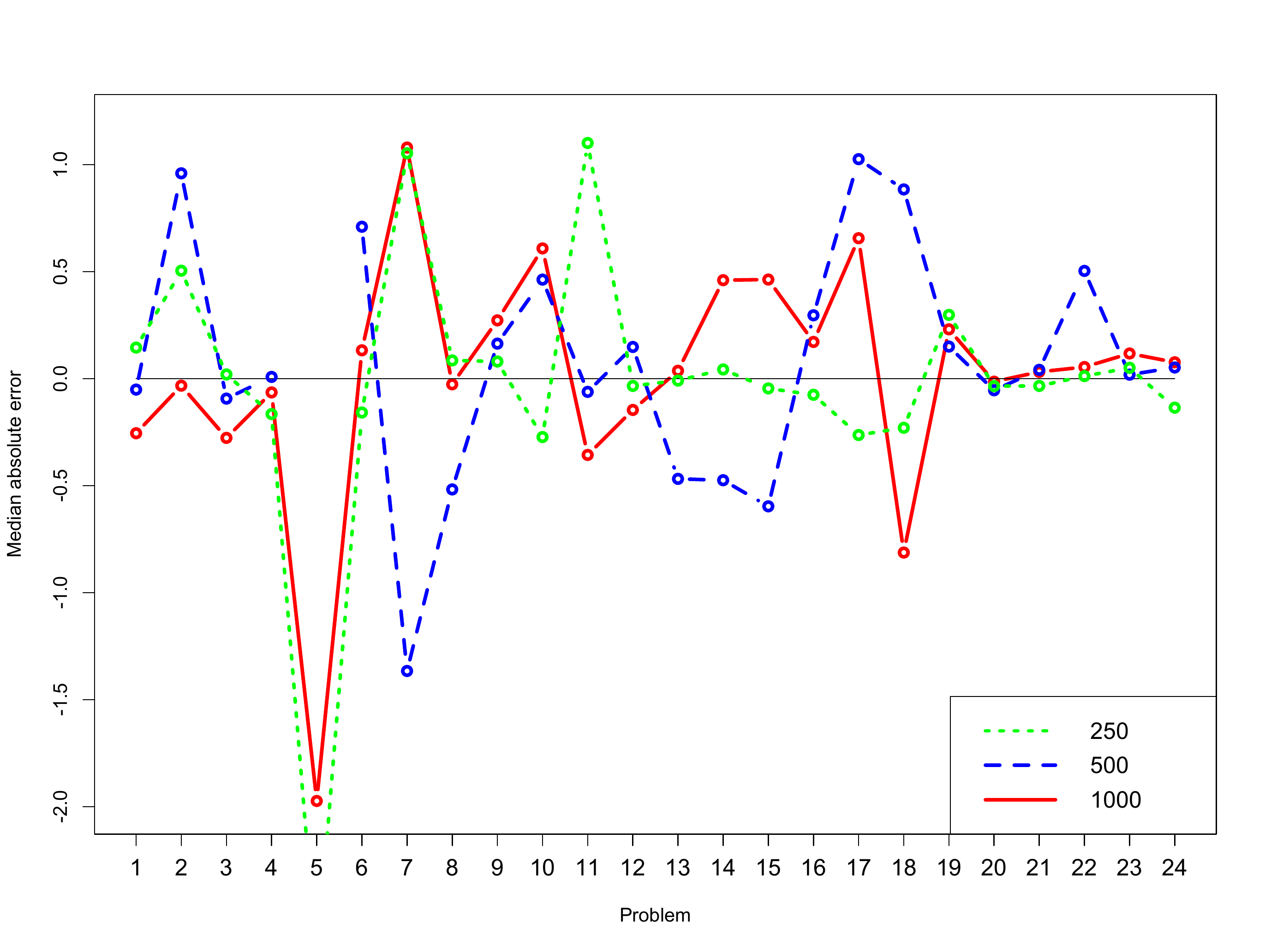}
    \caption{Relative advantage of the Ensemble-class vs. Best-train for each of the 3 budgets (250, 500, 1000). The y-axis presents the difference between the median absolute errors. Positive values indicate where the Ensemble-class model is better than the Best-train model.}
    \label{fig:budget}
\end{figure}

\textbf{Fixed ELA portfolio, one optimization algorithm, different budgets:}
Figure~\ref{fig:budget} presents the relative advantage of Ensemble-class model vs. Best-train model for each of the 3 budgets (250, 500, 1000), in the case when BIPOP-CMA-ES (i.e., the natural logarithm of the target precision) is targeted. Here, the ELA features portfolio is fixed and calculated using a $400d$ sample size. Positive values indicate where the Ensemble-class model is better than the Best-train model. From the figure, it follows that personalized ensembles work also for small budgets. Looking at problem 7, it seems that, based on its ELA representation, we can have a good prediction of the target precision reached after 250 and 1000 evaluations, but the Best-train model is better when the budget is 500. The limitation here is that the ELA-problem instance representation is static and the same for all budgets, and only the reached target is different. This means that the ELA representation does not cover information about the algorithm's behavior (i.e., which parts from the search space are visited until some budget). To improve this, further information about the state of the algorithm should be considered and used as input data to train the personalized ensembles for predicting performance in different budgets. 
\begin{figure}[!tbh]
  \setlength{\belowcaptionskip}{-10pt}
  \centering
  \subfloat[CMA-ES-CSA.]{\includegraphics[width=0.4\textwidth]{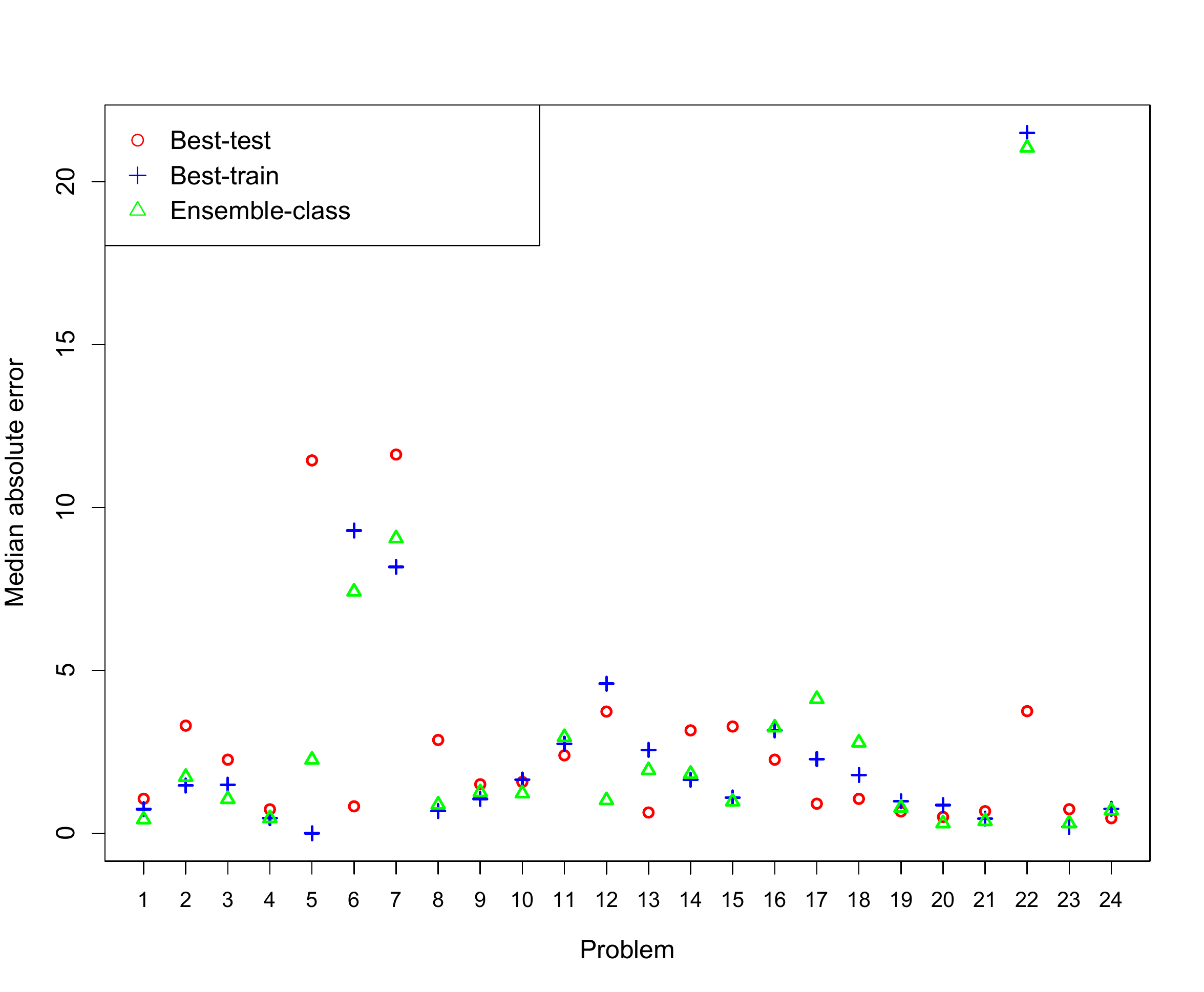}\label{fig:4alg}}
  \vfill
  \subfloat[IPOP400D.]{\includegraphics[width=0.4\textwidth]{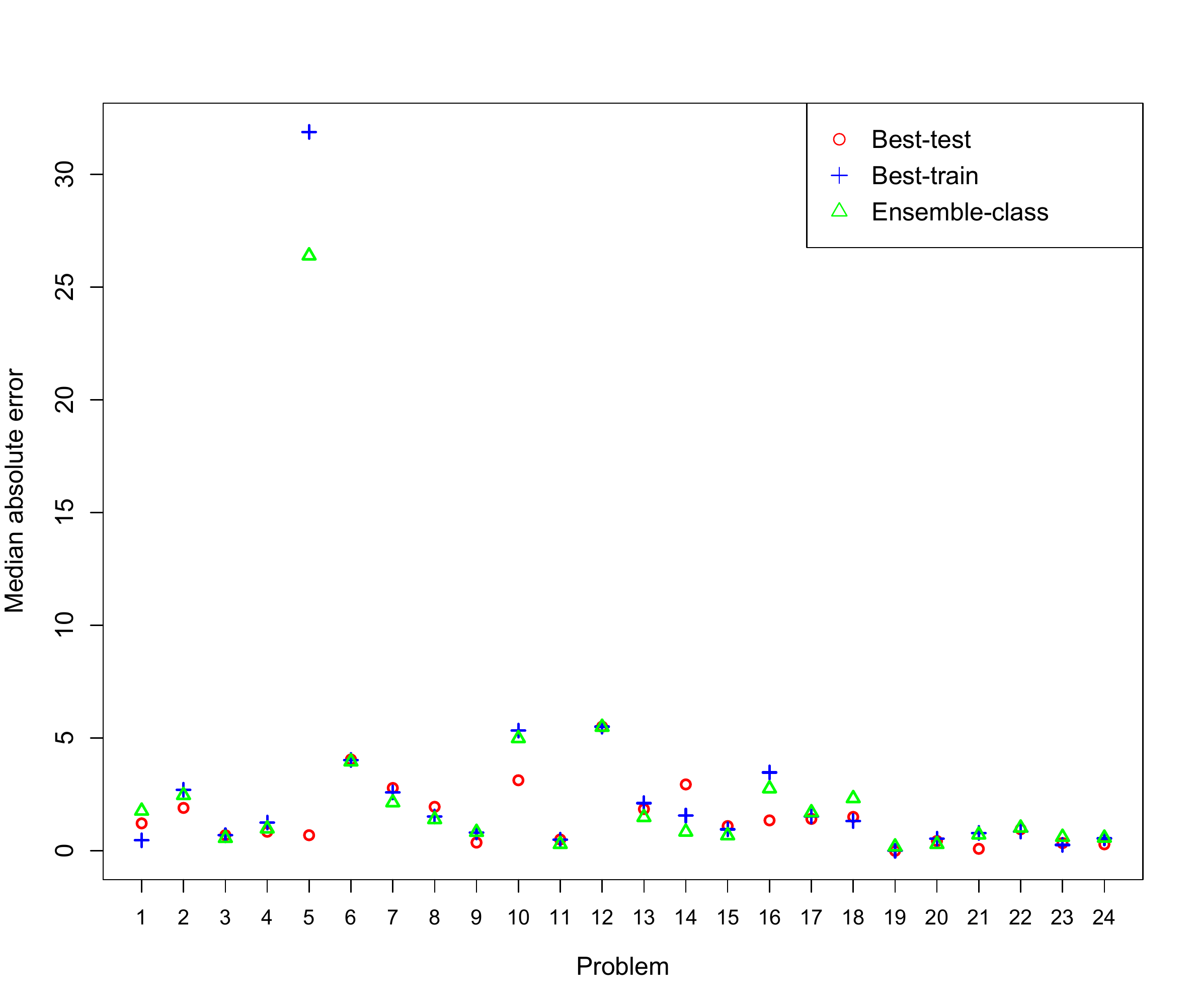}\label{fig:7alg}}
  \caption{Median absolute error between the truth and predicted target precision (i.e., natural logarithmic of the target precision) for each BBOB benchmark problem, for fixed budget 1000, fixed ELA features portfolio calculated using 400\textit{d} sample size, and two optimization algorithms.}
  \label{fig:opt-alg}
\end{figure}

\textbf{Fixed budget, fixed ELA portfolio, different optimization algorithms:}
To show the transferability of the proposed methodology to other algorithms than BIPOP-CMA-ES, personalized ensembles were learned to predict the performance of CMA-ES-CSA and IPOP400D, with a fixed budget of 1000 evaluations, and with a fixed ELA feature portfolio calculated using $400d$ sample size (Figure~\ref{fig:opt-alg}). Without going in detail into the results, it is obvious that using personalized ensembles improves the final prediction for most of the problems for both algorithms.

\section{Discussion and Future Works}
\label{sec:conclusions}

The paper presents the idea of predicting the performance of optimization algorithms, with the aim of selecting a regression model (or an ensemble) for a problem type. Our results demonstrate that there is quite some potential in moving from ``generalist'' regression models that work well across broad ranges of optimization problems to more problem-specific, personalized regression models. The sensitivity analyses confirm the robustness of our approach.  

We note that our study should be seen as a first prototype only. Several extensions are possible and needed. For example, we need to evaluate our methodology on much bigger data sets, to allow for a split into train, validation, and test sets. With such a setting, the train instances are used to train the RMs, the validation instances are then used to select and to evaluate the RMs to be included in the ensembles (this comprises the association of the importance weights that are used to calibrate the predictions of the different models). The test instances are then used to assess the performance of the overall pipeline. 

As far as the combination of the output of the different regression models into one prediction is concerned, we plan on evaluating different approaches to derive the weighting schemes. In particular, we believe that a multi-criteria approach to combine different regression performance measures (such as mean root square error, correlation coefficients, etc.) could be promising, to balance the complementary information obtained through each of these statistics. 

We used in this work multi-class classification  to assign problem instances to problem classes. In practice, instances may stem from problem classes not used in the training phase, so that the classifier cannot assign it to one of the present classes. In such cases, 
the classification step can be changed with clustering that will return its $k$ closest problem instances. Then, the personalized ensembles for the selected problem instances will be used to calculate the performance prediction, which will be further merged with some heuristic to generate the end prediction.

Last but not least, we will evaluate the personalized ensembles trained on one benchmark suite (the BBOB functions in our case) on other benchmark suites (e.g., Nevergrad~\cite{nevergrad}), in order to investigate the transferability of the models between the different benchmark collections. 

\vspace{1ex}
\textbf{Acknowledgments.} 
We thank Diederick Vermetten, Leiden University, for providing us the BBOB workshop data~\cite{BBOBdata} in a format that was conveniently post-processed by IOHanalyzer~\cite{IOHanalyzer}. We also thank Pascal Kerschke, University of Dresden, for providing us with the \texttt{flacco} tool~\cite{flacco} used to compute the feature values for the 24 BBOB functions. 

Our work was supported by projects from the Slovenian Research Agency (research core funding No. P2-0098, project No. Z2-1867, and grant number PR-10465), by the Paris Ile-de-France region, and by COST Action CA15140 ``Improving Applicability of Nature-Inspired optimization by Joining Theory and Practice (ImAppNIO)''. 
% \end{acks}

\bibliographystyle{ACM-Reference-Format}
\bibliography{references} 

\end{document}